\documentclass[letterpaper, 10pt, conference]{ieeeconf}      

\IEEEoverridecommandlockouts                              
\usepackage{graphics} 
\usepackage{graphicx, subfigure}
\usepackage{epsfig} 
\usepackage{mathptmx} 

\usepackage{amsmath}
\usepackage{amsfonts}
\usepackage{amssymb}
\usepackage{algorithm}
\usepackage{algorithmic}
\usepackage[T1]{fontenc}
\usepackage{color}
\usepackage{nomencl}
\usepackage[table]{xcolor}
\usepackage{algorithm}
\usepackage{listings}
\usepackage{float}
\floatstyle{ruled}
\newfloat{mylisting}{htbp}{loa}
\floatname{mylisting}{Listing}

\definecolor{propose}{rgb}{0.0,0.6,0.0}
\definecolor{warn}{rgb}{0.7,0.0,0.0}

\usepackage{listings}
\lstdefinelanguage{Lua}{%
morekeywords={%
and,break,do,else,elseif,end,false,for,function,if,in,%
local,nil,not,or,repeat,return,then,true,until,while%
},%
sensitive=true,%
morecomment=[l]{--},%
morecomment=[n]{--[[}{]]},%
morestring=[b]",%
morestring=[b]',%
}[keywords,comments,strings]%

\definecolor{comment}{RGB}{0,0,0}

\title{\LARGE \bf
Enhancing software module reusability using port plug-ins: an experiment with the iCub robot
}

\author{Ali Paikan, Vadim Tikhanoff, Giorgio Metta and Lorenzo Natale%
\thanks{This work was supported by the European FP7 ICT project No. 270273 (Xperience), project No. 288382 (POETICON++) and project No. 611832 (WALK-MAN).}
\thanks{A. Paikan, V. Tikhanoff, G. Metta and L. Natale are with the Istituto Italiano di Tecnologia (IIT), Genova, Italy. Emails: {\tt\small ali.paikan, vadim.tikhanoff, giorgio.metta, lorenzo.natale @iit.it }}%
}

\begin{document}

\maketitle
\thispagestyle{empty}
\pagestyle{empty}

\begin{abstract}
Systematically developing high--quality reusable software components is a difficult task and requires careful design to find a proper balance between potential reuse, functionalities and ease of implementation. Extendibility is an important property for software which helps to reduce cost of development and significantly boosts its reusability. This work introduces an approach to enhance components reusability by extending their functionalities using plug-ins at the level of the connection points (ports). Application--dependent functionalities such as data monitoring and arbitration can be implemented using a conventional scripting language and plugged into the ports of components. The main advantage of our approach is that it avoids to introduce application--dependent modifications to existing components, thus reducing development time and fostering the development of simpler and therefore more reusable components. Another advantage of our approach is that it reduces communication and deployment overheads as extra functionalities can be added without introducing additional modules.

The details of the plug-in system is described in the paper and its advantages for the development of robotics applications are demonstrated by developing a step--by--step example on the iCub humanoid robot. 
\end{abstract}

\section{Introduction}
Robotics software community is continuing to grow. Within the community, researchers have been developing large number of software components using some of the most common robotic middleware, such as ROS~\cite{quigley2009ros}, YARP~\cite{yarp07}, OROCOS~\cite{bruyninckx01orocos}, OPROS~\cite{jang2010opros} and Open-RTM~\cite{ando08} or based on their customized frameworks using standard communication libraries (e.g., CORBA~\cite{CORBA08}, ICE~\cite{ice-url}, \O{}MQ~\cite{zeromq}). They try to adopt lessons learned from best practices in robotics~\cite{Brugali2009, Brugali2010} and software architecture techniques and standards~\cite{Sametinger1997} to build their modules as reusable as possible. Even so, it is quite unlikely that components from different communities fit into a specific off--the--shelf deployment scenario, without any adaptation by third party users. Heterogeneity and lacking standards are not the only bottlenecks burdening reusability. Even within a community of developers who share the same middleware, software components can be developed with different taste and hard reuse. Systematically developing high--quality reusable software component is, indeed, a difficult task. Many developers keep their modules simple. However, simplicity does not necessarily lead to more reusable software. On the other hand, with reusability in mind, there is a risk of over--generalization and increased complexity: to build a more generic and reusable component, the developer tries to foresee all possible future needs and add them as reconfigurable functionalities to the software. Such a commitment leads to complex components, polluted with application--dependent functionalities that are more costly and difficult to maintain and use correctly. Thus, a proper balance must be found between potential reuse and ease of implementation~\cite{Sametinger1997}.

Software should be extensible enough to be adapted to possibly unanticipated changes~\cite{zenger2004programming}. Extensibility is an important property for software which significantly boosts reusability. One direction to extend a module is via its interfaces. In distributed systems interfaces are implemented by exchanging messages through special connection points that are call ports. This plays an important role in nowadays robotic software architectures. This paper concentrates on enhancing robotic software module's reusability by extending its port's functionality using a scripting language. The basic idea is to extend the port's functionalities in order to dynamically load a run--time script and plug it into the port of an existing module without changing the code or recompiling it. In our framework a port extension is called \emph{Port Monitor}: in brief it allows to access the data passing though a connection from/to the port for monitoring, filtering and transforming it (See Section~\ref{port-monitor}). Multiple port monitors can interact to allow an input port to select data from multiple sources in an exclusive way. We call this object a \emph{Port Arbitrator}: in other words, a port arbitrator allows a module to arbitrate data coming from other components to its input port and coordinate the corresponding modules (See Section~\ref{port-arbitrator}). {\color{comment}The detailed explanation of the port monitoring and arbitration system and their potential applications is given in \cite{paikan2014-portmonitor}.}  

The paper is structured as follow: Section~\ref{related-work} overviews the state of the art and related work. The detail of the Port Monitor and an overview of its API is described in Section~\ref{port-monitor}. Section~\ref{port-arbitrator} represents the Port Arbitrator. The applicability of our approach is demonstrated in Section~\ref{experiment} through a step--by-step example using the iCub robot's software repository. In Section~\ref{conclusion} we conclude the paper.

\section{Related Work}
\label{related-work}
Plug-in platforms, in general, extend a core system with new features implemented as components that are plugged into the core at run time and integrate seamlessly with it. When an application supports plug-ins, it enables customization, thus, provides a promising approach for building software systems which are extensible and customizable to the particular needs~\cite{Wolfinger-2006}. Probably one of the more prominent example of a platform which broadly supports plug-ins is Eclipse IDE~\cite{gamma2004contributing}. Eclipse offers a framework to develop plug-ins in Java which are delivered as JAR libraries. There are also some generic frameworks for plug-in development and management such as Pluma~\cite{pluma} which allows loading plug-ins as dynamic linked libraries or FxEngine~\cite{fxengine} for data flow processing and the design of dynamic systems. Plug-ins can also be developed using scripting languages. Scripting languages have been used for decades to extend the functionality offered by software components and they have special interests within the game developer communities. The main advantage of script--based plug-ins is that they are usually easier to be developed and maintained.  

Despite plug-in system has been broadly used by software developers over the last decades, to our knowledge, less attention has been devoted to study their potentials in the robotic field. The work presented in this paper is an approach to extend a YARP component's port as it can act as a plug-in manager to load plug-ins written in a scripting language~\footnote{The source code and relevant examples can be found at {https://github.com/robotology/yarp/tree/master/src/carriers/}}. The current implementation offers a plug-in development using Lua but it can be easily extended to support other languages. 

\section{Port Monitor}
\label{port-monitor}
\begin{figure}[t]
   \begin{center}
     \includegraphics[width=2.5in]{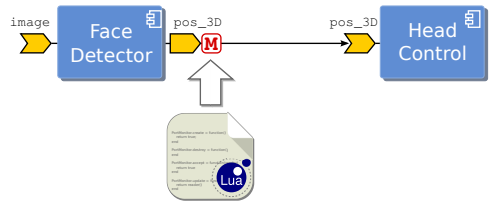}
     \caption{Conceptual representation of port monitor. The output port of Face--Detector modules is extended with a plug-in which provides access to the outgoing data through scripting language (e.g., Lua).}
     \label{fig:port-monitor}
   \end{center}
 \end{figure}

To better illustrate the concept of the Port Monitor, we can consider, as an example, an application for tracking faces in a humanoid robot as shown in Fig.~\ref{fig:port-monitor}. The application involves two simple modules. The first one is a Face--Detector module which receives image data from the robot's camera, detects human faces and streams out, through its output port, the 3D position of the detected face together with a confidence level. The second module is called Head--Control; which in its turn receives a 3D position and controls the head of the robot to look at the corresponding point. A simple head tracking application can be achieved by connecting the output of the Face--Detector to the input of the Head--Control. Now suppose we want to extend the application and track the face only if the confidence level is above a certain threshold. This can be achieved by modifying the Head--Control module to take into account the extra information, but then it would be inadvertently polluted with application--dependent functionality. A more appropriate choice would be to develop a third module which receives data from the Face--Detector and filters out messages corresponding to detections that do not satisfy the required confidence level. The drawback of this approach is that it introduces extra implementation effort whilst adding further communication and deployment overhead to the system. 
 
One could argue over the immaturity of the involved modules and proposes that, for example, the Face--Detector module could be improved and reimplemented so that it can be reconfigured by specifying the desired confidence level. This solution requires higher development cost and in general it easily leads to complex design. In addition, it makes it even more arduous to connect the Face--Detector to multiple modules with different confidence requirements, thus preventing runtime reusability. 

In our approach it is possible to solve this problem by extending the ports of components using run--time scripts. Fig.~\ref{fig:port-monitor} represents the concept of the Port Monitor (shown as a box with M) attached to the output of the Face--Detector module. The Port Monitor can load a script file (written using a standard scripting language such as, in our case, Lua~\cite{ierusalimschy1996lua}) and can access and modify the data traveling through the port using a simple API. Thus, some extra functionalities of a component such as data filtering, transformation, monitoring can be added during the application development time and without the need to modify and rebuild the component itself. Similarly the same script can be loaded by the input port of the Head--Control module. 

 \begin{figure}[t]
   \begin{center}
     \includegraphics[width=2.5in]{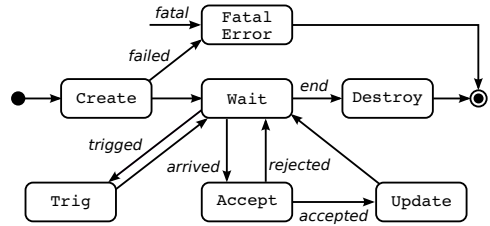}
     \caption{The life cycle of port monitor.}
     \label{fig:port-monitor-lifecycle}
   \end{center}
 \end{figure}
 
This mechanisms offers the following advantages. Firstly, it avoids adding to the component application--specific functionalities. That is, some application--dependent functionalities can be freely added to the component using the scripting language during the application development stage. Secondly, it allows to simplify the implementation of the components, since the developer does not necessarily need to provide all possible configurations supporting different application scenarios. Finally, by embedding the extra functionalities inside the port, our approach intrinsically reduces communication and deployment overhead that would be  introduced if the same functionalities were added as separate modules.

\subsection{Port monitor life cycle and API}

Fig.~\ref{fig:port-monitor-lifecycle} illustrates the states that define the life cycle of a port monitor. A callback function is assigned to each state (except Waiting) which can have a corresponding implementation in the user's script. Using these callbacks, users have full control over the port's data and can access it, modify it and decide whether to accept the data or discard it. Listing~\ref{lst:callback} represents the callback functions corresponding to the port monitor's states in Lua.  

The Port Monitor starts in the \emph{Create} state in which \texttt{PortMonitor.create} callback is called. The initialization of the user's code can be done at this point. Returning a true value means that the user's initialization was successful and the monitor object is able to start monitoring data from the port. When data arrives to the monitor, \texttt{PortMonitor.accept} is called. In this callback, user can access (for reading purposes only) the data, check it and decide whether to accept or discard it. The return value of this function indicates whether the data should be delivered (accepted) or discarded. If the data is accepted, \texttt{PortMonitor.update} is called, at which point the user has access to {\it modify} the data.    
 
\lstset{captionpos=b, caption={Port monitor callback functions in Lua},label=lst:callback}
\begin{lstlisting}[language=Lua, frame=single]
 PortMonitor.create = function() return true end
 PortMonitor.accept = function(dt) return true end
 PortMonitor.update = function(dt) return dt end  
 PortMonitor.trig = function() return end
 PortMonitor.destroy = function() end
\end{lstlisting}

A port monitor will usually act as a passive object~\cite{Kenichiand1995} where accept and update callbacks are called only upon data reception. However, one may need to periodically monitor a connection (within a specific time interval) and, for example, generate proper events in the case of delay in the communication. For this purpose, a port monitor object can be configured to call \texttt{PortMonitor.trig} within desired time intervals. Finally, \texttt{PortMonitor.destroy} is called when the port monitor is detached from the port upon disconnection. As an example listing~\ref{lst:data_monitor} illustrates the pseudo--script in Lua that in the hypothetical application that requires filtering out messages from Face--Detector when the confidence level is below a threshold of $80 \%$.   

\lstset{%
label=lst:data_monitor,
caption={An example of filtering Face--Detector data.},
frame=single,
captionpos=b,
numbers=left,
numbersep=-8pt,
framexleftmargin=-5pt,
framexrightmargin=-5pt,
numberstyle=\tiny\color{darkgray}
}
\begin{lstlisting}[language=Lua]
   PortMonitor.accept = function(data)
      -- read face_pos from `data'
      if face_pos.certainty < 0.8 then	
         return false
      end      
      return true 
   end
\end{lstlisting} 

\section{Port Arbitrator}
\label{port-arbitrator}
\begin{figure}[t]
  \begin{center}
    \includegraphics[width=2.5in]{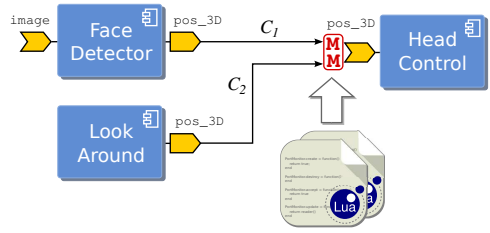}
    \caption{Conceptual representation of port arbitrator}
    \label{fig:port-arbitrator}
  \end{center}
\end{figure}

A port Arbitrator is an extended functionality of an \emph{input} port which can be configured to arbitrate data from multiple source based on some user--defined constraints. Fig.~\ref{fig:port-arbitrator} represents a simple search--and--track application where a humanoid robot looks around in search of a person's face and tracks it. The robot should look around only if it is not tracking a face. The application involves modules described in the face--tracking example from Section~\ref{port-monitor} and an extra module to look around. Look--Around generates some random 3D position which makes the robot randomly look around when it is provided to the Head--Control module. To obtain the desired behavior, the Head--Control module should \textbf{not} receive data from the Look--Around module when the Face detector module is sending detected face positions.

This is a common coordination problem which can be solved in different ways (e.g., using a separate coordinator, extending modules to interact with each other). One way to achieve this is to use a selector in the input port of the Head--Control module and constrain it to receive data from each module under specific conditions. The concept is shown in Fig.~\ref{fig:port-arbitrator} where a port arbitrator is used in the input port of the Head--Control (shown as box labeled with two 'M'). The arbitration logic can be written using a scripting language and is loaded by the port arbitrator. Our previous work~\cite{paikan13} has demonstrated that this type of arbitration mechanism can be effectively used to implement complex tasks without resorting to centralized coordinators. 
      
\subsection{Architecture of Port Arbitrator}
\begin{figure}[t]
    \begin{center}
      \includegraphics[width=2.1in]{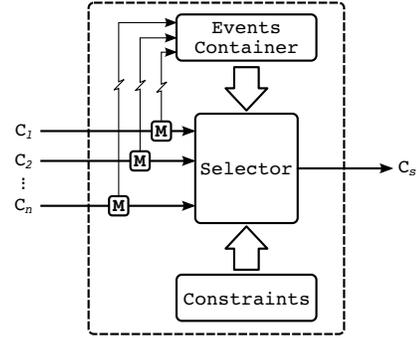}
      \caption{The architecture of port arbitrator. Straight lines show the data flow and zigzag lines represent event flows.}
      \label{fig:arbitraion_model}
    \end{center}
\end{figure}

Fig.~\ref{fig:arbitraion_model} represents the internal architecture of the port arbitrator. A port arbitrator consists of multiple port monitors, a set of selection constraints, an event container and a selector block. Port arbitrator extends the port's scripting API for setting constraints and altering events in the container. In fact, when a port monitor is used in an arbitrator, the user's script can access the extended API for arbitration. 
 
A port monitor can be attached to each connection ($C_i$) going through the port arbitrator. It monitors the data from connection and inserts the corresponding events into a shared container. A port monitor can also remove an event (if previously inserted by itself) from the container~\footnote{This is similar to the Event--Mask mechanism used in user interface programming or in operating systems.}. Normally events have infinite life time. This means that they remain valid in the container until they are explicitly removed by the monitor object. An event can also have a specific life time. A time event will be automatically removed from the container when its life time is over. For each connection $C_i$, there is a selection constraint written in first order logic as a boolean combination of the symbolic events. Upon the reception of data from a connection, the selector evaluates the corresponding constraint and, if satisfied, it allows the data to be delivered to the input port; otherwise the data will be discarded. Clearly a consistency check on the boolean rules must be performed to guarantee that only a single connection $C_i$ can deliver data at any given time. Listing~\ref{lst:arbitrator-api} represents the extended port monitor's API in Lua which can be used with port arbitrator. 

\lstset{captionpos=b, numbers=none,%
 caption={Port monitor extended API in Lua for arbitration}, 
 label=lst:arbitrator-api}
\begin{lstlisting}[language=Lua, frame=single]
 PortMonitor.setEvent(event, life_time)
 PortMonitor.unsetEvent(event)
 PortMonitor.setConstraint(rule)
\end{lstlisting}

We refer to the search--and--track example from Fig.~\ref{fig:port-arbitrator} to demonstrate how selection constraints are represented and how they can be evaluated based on events from a container. As we previously mentioned, the Head--Control module should receive data from the Look--Around module if the Face--Detector module is not sending any data. To do this, we first need to inform the port arbitrator about the status of the data from the Face--Detector (i.e., if it is sending any data or not) by setting an event into the container.  Listing~\ref{lst:arbitrator-setevent} represent a simple script to set the \texttt{`e\_face\_detected'} into the event container. The life time $1.0$ indicates that \texttt{`e\_face\_detected'} will be automatically removed after one second if the Face--Detector is not sending any data.  
 
\lstset{%
label=lst:arbitrator-setevent,
caption={An example of setting a time event into a container.},
frame=single,
captionpos=b,
numbers=left,
numbersep=-8pt,
framexleftmargin=-5pt,
framexrightmargin=-5pt,
numberstyle=\tiny\color{darkgray}
}
\begin{lstlisting}[language=Lua]
   PortMonitor.accept = function(data)
      setEvent("e_face_detected", 1.0)
      return true 
   end
\end{lstlisting} 

At this stage, the selection rule that allows the data from the Look--Around module ($C_{2}$) to be delivered to the Head--Control module when \texttt{`e\_face\_detected'} does not exist in the event container, can be simply written as follows: \texttt{setConstraint(`not e\_face\_detected')}. The data from the Face--Detector module should be freely delivered to the Head--Control. Thus the selection rule for the connection $C_{1}$ is written as follows: \texttt{setConstraint(`true')}.  As we previously described, constraints can be expressed as boolean combinations of symbolic events. To evaluate the expression, every symbolic event is substituted with a boolean value {\color{comment} which represents its logical state}. If the event is present in the container, it represents a true value in the expression; otherwise it is evaluated as false.   

\begin{figure}[t]
   \begin{center}
     \includegraphics[width=2.5in]{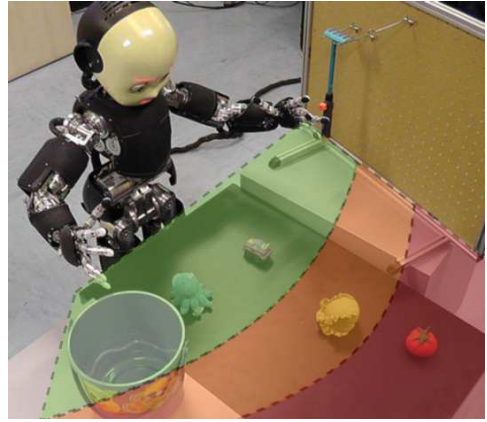}
     \caption{The experimental setup of table--cleaning application. The reachable zone is depicted in green, the orange zone represents the zone reachable with the tool and finally the red zone indicates the unreachable space, for which the robot needs human intervention. }
     \label{fig:experiment-setup}
   \end{center}
\end{figure}

\begin{figure}[t]
   \begin{center}
     \includegraphics[width=2.5in]{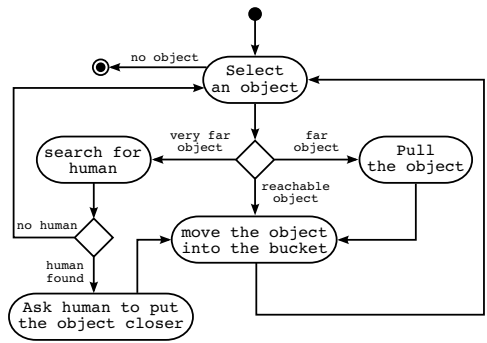}
     \caption{The simplified activity diagram that illustrate the table--cleaning application.}
     \label{fig:experiment-activity}
   \end{center}
\end{figure}
 
\section{A step--by-step example}
\label{experiment}
To demonstrate the applicability and advantages of our approach, we present an experiment\footnote{The video which demonstrates the complete experiment can be accessed at http://www.youtube.com/user/robotcub.} with the iCub humanoid robot~\cite{Metta2008}. This experiment is completely built using modules from the iCub software repository\footnote{Modules can be downloaded from: https://github.com/robotology/icub-main.git and https://svn.code.sf.net/p/robotcub/code/trunk/iCub/contrib}. %
The experiment focuses on reusing (with no modifications) existing modules and by extending the required functionalities using port plug-ins. The overall behavior of the experimental task is demonstrated using a simplified activity diagram in Fig.~\ref{fig:experiment-activity}. The goal of the task, as shown in the activity diagram, is to clean the table by removing all the object and place them in a bucket located alongside the table. We allowed the robot to use a tool at his disposal (a rake), located on a rack, to reach objects of interest that are out of his workspace. The modules that allow the robot to grasp and use the tool are implemented as described in~\cite{Tikhanoff-2013}. Furthermore, we consider also the case in which the object is so far that it cannot be reached even by the use of the tool. In this case the robot should look for a human and asks his intervention (put the object within reach). Fig.~\ref{fig:experiment-setup} shows the experimental setup and it illustrates the three areas in which objects can be placed. 

The activity diagram depicted in Fig.~\ref{fig:experiment-activity} may give the impression that the task is only composed of a few simple steps that the robot should follow to accomplish it. But in fact, there are many uncertainties and unexpected conditions which should be taken into consideration to make the task robust. For example, the proper decision should be taken if an object drops from the hand while the robot is placing it into the bucket. Similarly the robot should behave appropriately while it is holding the tool to pull the object closer, the human might intentionally intervene and move the object within the iCub's workspace. Considering all possible uncertainties, in fact, reveals the underlying complexity of the task which requires that many modules (e.g, for perception, action and coordination) are properly used and orchestrated (e.g, coordinating robots, gaze, arm, speech) to perform the required task.

\begin{table}[ht]
\caption{A subset of modules used for the experiment}
\centering
\begin{tabular}{l c c c} 
\hline\hline 
Module & Input & Output & Type\\ [0.5ex]
\hline
\textbf{Face--Detector} & image & pos\_3D & perception\\
\textbf{Object--Detector} & image & List<pos\_3D> & perception\\
\textbf{Bucket--Detector} & image & pos\_3D & perception\\
\textbf{Look--Around} & - & pos\_3D & implicit action\\
\textbf{Head--Control} & pos\_3D & - & action\\
\textbf{Pick--and--Place} & msg\_cmd & msg\_status & action\\
\textbf{Pull--Object} & msg\_cmd & msg\_status & action\\
\textbf{Speak} & msg\_text & - & action\\
\hline
\end{tabular}
\label{tb:modules} 
\end{table}

The modules used in this experiment are chosen from the iCub software repository and listed here in Table~\ref{tb:modules}. To build the desired application, a few modules might simultaneously require to grab the camera image frames from the robot, control the arms and hands in various modes, such as Cartesian or joint space using velocity or position control. However, for the sake of brevity, only a subset of these modules are described in this paper. We use the previously mentioned Face--Detector and the Look--Around modules. 

Object--Detector gets as an input image from the cameras and produces a list of blobs and extracts 3D positions of all the possible graspable objects as its output. Bucket--Detector is, in fact, an instance of a generic object detector which is configured and trained to recognize this specific object. As we previously mentioned, Look--Around randomly produces positions in 3D space which are used by Head--Control to move the gaze in various positions. The Pick--and--Place module receives a set of commands (e.g., \texttt{take <3D\_pos>}, \texttt{put <3D\_pos>}) to take an object and release it on a specific position. The internal status of the module (e.g., \texttt{e\_taken}, \texttt{e\_arm\_idle}) is continuously sent out using status messages. Pull--Object is a complex set of modules which together get the position of an object on the table and use a tool to bring the object closer~\cite{Tikhanoff-2013}. Similar to Pick--and--Place, the internal status of the Pull--Object module is advertised via its output. The Speak module receives a text message and performs a text--to--speech synthesis. Generally speaking, in order to be able to integrate some modules for building an application, two important points should be considered: \emph{i)} data type on both side of the connections should match and \emph{ii)} a proper coordination mechanism should orchestrate modules to perform the task. We start with the simplest case in which the objects are reachable by the iCub and progressively extend it to build the complete table--cleaning application. 

\subsection{Handling reachable objects}
\begin{figure}[t]
   \begin{center}
     \includegraphics[width=2.2in]{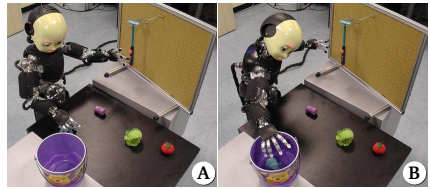}
     \caption{The iCub performing table--cleaning on reachable objects. The robot takes the object (A) and places it into to bucket (B).}
     \label{fig:exp-reachable}
   \end{center}
\end{figure}

First our application should select the closest object within the reachable area and take it (see Fig.~\ref{fig:exp-reachable}-A ). To do that, we connect the output of Object--Detector to the input of Pick--and--Place. Using the port monitor, we implement a simple script that goes through the list of objects, select the one that is closest to the robot and produces the proper `take' command (i.e., \texttt{take <3D\_pos>}) for execution. Similarly, to put the object into the bucket we connect the output of Bucket--Detector with the same input of Pick--and--Place and attach to this connection another port monitor that generates the `put' command (i.e., \texttt{put <3D\_pos>}) for execution (see Fig.~\ref{fig:exp-reachable}-B )

\begin{figure}[t]
   \begin{center}
     \includegraphics[width=2.5in]{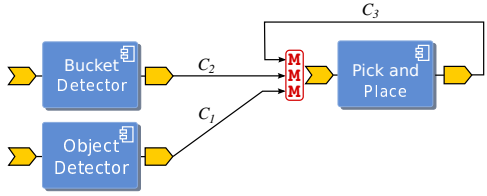}
     \caption{Configuration of the modules for handling reachable objects on the table.}
     \label{fig:close-objects}
   \end{center}
\end{figure}

Furthermore, an object should be taken only if the hand of the robot is free and the robot is not performing another action using the arm. On the other hand, the `put' command should be sent to the Pick--and--Place module if the robot is holding an object. To this aim, the status of the Pick--and--Place module should be monitored and the required arbitration rules should be added to the system to properly coordinate taking, placing and releasing actions. Fig.~\ref{fig:close-objects} represents the configuration of the modules that perform this simple task on the reachable objects. As shown in the figure, the status output of the Pick--and--Place module is used to inform the arbitrator about the internal state of the module. Below we illustrate how this is achieved.


As we have mentioned previously, a monitor object is assigned to each connection going through the port arbitrator. Listings~\ref{lst:close-object-c1},~\ref{lst:close-object-c2} and~\ref{lst:close-object-c3} respectively represent pseudo--scripts which will be loaded by each monitor object for connections $C_1$, $C_2$, and $C_3$. Listing~\ref{lst:close-object-c3} demonstrates the script which is assigned to the monitor object of connection $C_3$. This monitor receives status messages from Pick--and--Place (i.e., \texttt{e\_taken}, \texttt{e\_arm\_idle}) and adds them to the event container of the port arbitrator. These events will be used for the selection of $C_1$ and $C_2$. Notice that the connection $C_3$ and the corresponding script (Listing~\ref{lst:close-object-c3}) are created to make the status events available for the arbitration. These events will be never delivered to Pick--and--Place. This is achieved by refusing to accept the data from the connection $C_3$ (\texttt{return false}).

\lstset{%
label=lst:close-object-c1,
caption={Monitoring and arbitrating connection $C_1$.},
frame=single,
captionpos=b,
numbers=left,
numbersep=-8pt,
framexleftmargin=-5pt,
framexrightmargin=-5pt,
numberstyle=\tiny\color{darkgray}
}
\begin{lstlisting}[language=Lua]
   PortMonitor.create = function()
       setConstraint("not e_taken and e_arm_idle")
      return true;
   end
        
   PortMonitor.accept = function(object_list)
         -- find closest_obj in the object_list
         if closest_obj.dist > HAND_REACHABLE then	
            return false
         end         
      return true 
   end

   PortMonitor.update = function(object_list)
      return command("take", closest_obj.pos)
   end
   
\end{lstlisting} 

\lstset{%
label=lst:close-object-c2,
caption={Monitoring and arbitrating connection $C_2$.},
frame=single,
captionpos=b,
numbers=left,
numbersep=-8pt,
framexleftmargin=-5pt,
framexrightmargin=-5pt,
numberstyle=\tiny\color{darkgray}
}
\begin{lstlisting}[language=Lua]
   PortMonitor.create = function()
      setConstraint("e_taken and e_arm_idle")
      return true;
   end

   PortMonitor.update = function(bucket_pos)
      return command("put", bucket_pos)
   end   
\end{lstlisting}

\lstset{%
label=lst:close-object-c3,
caption={Monitoring connection $C_3$ for generating events.},
frame=single,
captionpos=b,
numbers=left,
numbersep=-8pt,
framexleftmargin=-5pt,
framexrightmargin=-5pt,
numberstyle=\tiny\color{darkgray}
}
\begin{lstlisting}[language=Lua]
   PortMonitor.accept = function(status_event)
      setEvent(status_event, 0.5)
      return false 
   end   
\end{lstlisting}  

Listing~\ref{lst:close-object-c1} deserves particular attention: First, within the `create' callback, the required selection rule for the connection $C_1$ is set into arbitrator. The rule implies that data from corresponding connection should be delivered if the robot has not already taken (\texttt{not e\_taken}) an object and if it is not performing an action (\texttt{e\_arm\_idle}). In the `accept' callback, first the closest object to the robot is selected from the list of detected objects. If the object is reachable (the data is accepted), the `update' method will be called to generate the `take' message to be delivered to Pick--and--Place. If the object is out of reach, it will be discarded (\texttt{return false}).
{\color{comment} Similarly, Listing~\ref{lst:close-object-c2} represents the script to generate the `put' command and it also specifies the constraints for performing the corresponding actions.} 

\begin{figure}[t]
   \begin{center}
     \includegraphics[width=3.3in]{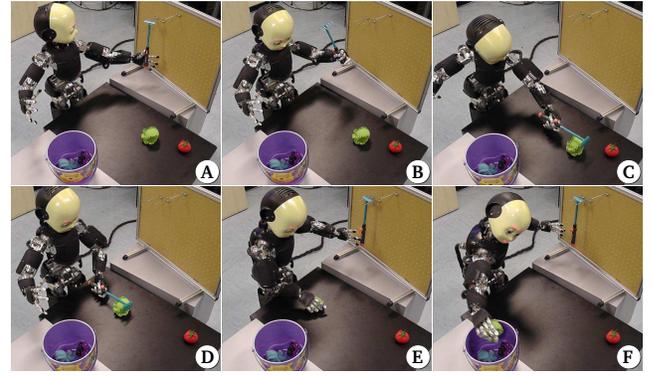}
     \caption{The iCub performing table--cleaning using a tool (rake). The robot take the tool (A), reaches for the object (B,C), pulls the object (D), grasps the object (E) and finally places it into the bucket (F).}
     \label{fig:exp-tool}
   \end{center}
\end{figure}

\begin{figure}[t]
   \begin{center}
     \includegraphics[width=2.5in]{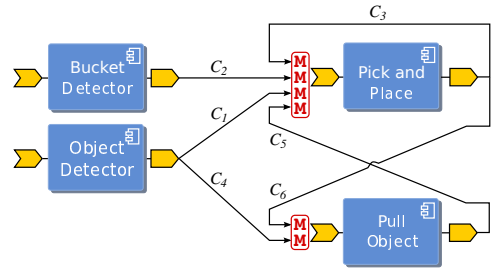}
     \caption{Configuration of the modules for handling objects within tool--reach space.}
     \label{fig:far-objects}
   \end{center}
\end{figure}

\lstset{%
label=lst:far-object-c4,
caption={Monitoring and arbitrating connection $C_4$.},
frame=single,
captionpos=b,
numbers=left,
numbersep=-8pt,
framexleftmargin=-5pt,
framexrightmargin=-5pt,
numberstyle=\tiny\color{darkgray}
}
\begin{lstlisting}[language=Lua]
   PortMonitor.create = function()
      setConstraint("not e_taken and e_arm_idle")
      return true;
   end
        
   PortMonitor.accept = function(object_list)
      -- find closest_obj in the object_list
      if closest_obj.dist > TOOL_REACHABLE then	
         return false
      end         
      return true 
   end

   PortMonitor.update = function(object_list)
      if closest_obj.dist < HAND_REACHABLE then
         return command("cancel", nill)
      end 
      return command("pull", closest_obj.pos)
   end   
\end{lstlisting} 

\subsection{Handling objects using tool}
Now, we extend the previous application to allow the iCub to use a tool to bring unreachable object within its workspace (see Fig.~\ref{fig:exp-tool} ). Fig.~\ref{fig:far-objects} represents how Pull--Object is integrated in the application. The output of Object--Detector module provides a list of objects; this list should be filtered to select one object that is within the tool--reach area and out of the robot's workspace. The position of this object should be given to the Pull-Object to trigger a sequence of actions to take the tool from the rack, reach for the object with the tool, pull the object and finally putting back the tool on the rack (see Fig.~\ref{fig:exp-tool}-B, C, D ). Once the object is located within the reachable area of the robot, the previous picking--and--placing application is activated. Appropriate selection rules should be added to the system to properly arbitrate pulling and pick--and--placing.%
\begin{figure}[t]
   \begin{center}
     \includegraphics[width=3.3in]{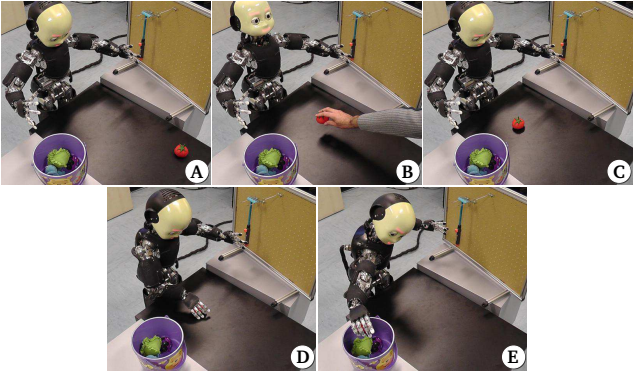}
     \caption{The iCub performing table--cleaning with human assistance. The robot detects an unreachable object (A), detects the presence of a human and asks assistance (B,C), grasp the object (D) and finally places it into the bucket (E). }
     \label{fig:exp-human}
   \end{center}
\end{figure}
Listing~\ref{lst:far-object-c4} represents the pseudo code of the script which is used in the port monitor of connection $C_4$. The selection constraint (\texttt{not e\_taken and e\_arm\_idle}) filters messages to Pull--Object when the robot is already involved in other actions (i.e. picking and placing an object). Similar to Listing~\ref{lst:close-object-c1} from the previous application, first the closest object is extracted from the list of detected objects. This object is accepted and generates a `pull' command if it is within the tool--reach area. Otherwise it is discarded. An interesting behavior is the fact that the pulling action is composed of several sub--actions that should be aborted if the tool becomes unnecessary (e.g. if a human moves the target objects in the workspace of the robot). This is achieved by continuously monitoring the target object in the `update' function and generating the 'cancel' command when necessary. Notice that as opposed to Pick--and--Place, Pull--Object ignores redundant `pull' commands until all ongoing sub-actions are accomplished or aborted (with the `cancel' command). Therefore, unlike Pick--and--Place, we do not need to monitor the internal status of Pull--Object and filter conflicting `pull' commands. %

Clearly Pick--and--Place and Pull--Object are conflicting behaviors. To avoid conflicts the selection rule for connection $C_1$ must be updated to prevent generation of `take' commands while Pull--Object is active (i.e. not idle). This is achieved by making the internal state of the Pull--Object available in the arbitrator of Pick--and--Place via connection $C_5$ and by modifying the selection constraint of Listing~\ref{lst:close-object-c1} as follows: 
\texttt{`not e\_taken and e\_arm\_idle and e\_pull\_idle'}. As for the connection $C_3$, Listing~\ref{lst:close-object-c3} is used for the port monitor of connections $C_5$ and $C_6$ to inserts the status events into the corresponding event containers. 
 
\subsection{Handling objects with human assistance}
\begin{figure}[t]
   \begin{center}
     \includegraphics[width=2.7in]{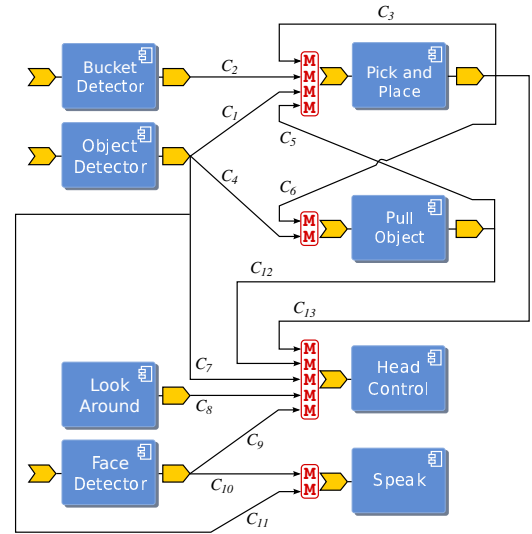}
     \caption{Configuration of the modules for table--cleaning application.}
     \label{fig:assist-objects}
   \end{center}
\end{figure}
In Section~\ref{port-arbitrator} we explained how the Face--Detector and Look--Around modules can be properly used with the Head--Control module to implement a basic face tracking application. In this section, we use these modules to complete our table--cleaning application. When an object is completely unreachable, the robot should look for a person and asks assistance (see Fig.~\ref{fig:exp-human}). Fig.~\ref{fig:assist-objects} depicts the complete system. The output of Object--Detector arbitrates the connections from Face-Detector and Look-Around via $C_7$ and $C_{11}$ so that when required, the robot will look around searching and tracking human faces.%
\lstset{%
label=lst:assist-c7,
caption={Monitoring connection $C_7$ and $C_{11}$.},
frame=single,
captionpos=b,
numbers=left,
numbersep=-8pt,
framexleftmargin=-5pt,
framexrightmargin=-5pt,
numberstyle=\tiny\color{darkgray}
}
\begin{lstlisting}[language=Lua]        
   PortMonitor.accept = function(object_list)
      -- find closest_obj in the object_list
      if closest_obj.dist > TOOL_REACHABLE then	
         setEvent("e_unreachable")
      else
         unsetEvent("e_unreachable")
      end         
      return false
   end
\end{lstlisting}
 
\lstset{%
label=lst:lst:assist-c10,
caption={Monitoring and arbitrating connection $C_{10}$.},
frame=single,
captionpos=b,
numbers=left,
numbersep=-8pt,
framexleftmargin=-5pt,
framexrightmargin=-5pt,
numberstyle=\tiny\color{darkgray}
}
\begin{lstlisting}[language=Lua]
   PortMonitor.create = function()
      setConstraint("e_unreachable")
      return true;
   end
        
   PortMonitor.accept = function(data)
      if time() - time_prev < DESIRED_TIME then
         return false  
      end
      time_prev = time()
      return true 
   end

   PortMonitor.update = function(data)
      return msg("Please put the object closer!")
   end
   
\end{lstlisting} 
This is achieved in Listing~\ref{lst:assist-c7} by monitoring the closest object and generating an event \texttt{'e\_unreachable'} when the latter is out of the tool--reach area. Notice that this event is cleared (removed from the container) only when the object becomes reachable again. Messages from Look--Around and Face-Detector are discarded depending on the internal state of Pick--and--Place and Pull--Object via connections $C_{12}$ and $C_{13}$ and the event generator script (i.e., Listings~\ref{lst:close-object-c3}). This prevents moving the head when the robot is picking, placing or attempting to pull an object. Finally the output of Face--Detector generates a voice message synthesized by the Speak module. This is achieved by connecting the two modules ($C_{10}$) and adding a script to the corresponding port monitor. This script generates a text message (a valid command for the Speak module) if a human face is detected, but only if a certain amount of time has passed from the last command, to reduce verbosity (Listing~\ref{lst:lst:assist-c10}). Notice that these commands are arbitrated by $C_{11}$ so that the speech is activated only when necessary.

\addtolength{\textheight}{-3cm}

\section{Conclusions and Future works}
\label{conclusion}
This article has introduced an approach that enhances software components reusability using port plug-ins. The key idea of our approach is to extend modules functionalities by adding scripts to the ports that allow to monitor, filter and transforming data and generating events. Another important concept is `port arbitration'; this allows to enhance the port capability by adding rules to arbitrate input data from multiple sources. The main advantage of our approach is that it allows to limit application specific functionalities to scripts that are external to the modules and are added and executed at runtime. This maintains modules clean from unnecessary complexity and enhances their reusability. Finally, by using embedded scripts inside the ports, we can avoid introducing specific modules to achieve the required functionalities, thus, reducing communication and deployment overhead. To demonstrate the potential advantages of our approach, we illustrated the implementation of a complex application on the iCub humanoids robot which was completely built out of existing modules without code changes. All the functionality specific to the application were implemented and integrated as plug-ins scripts.%

{\color{comment} The port plug--ins can be implemented in other distributed frameworks, as for example by extending existing connection ports with the functionalities required to load code and execute it to parse incoming or outgoing data. This extension can be easily implemented if the framework provides a callback mechanism or by actively monitoring a port using a dedicated thread. %
Perhaps one of the limitation of port plug--in is that it cannot be easily used in a service--oriented system where coordination of subsystems required interactions of components using bidirectional communication and remote procedure calls. However, port--arbitrated coordination does not constrain the application designer to a specific mechanism for the orchestration of components. For example, a subset of components can be coordinated using state machines and another subset using port arbitration.

In its current implementation, port arbitration requires that constraints are set via scripts and manually checked for consistency in the design phase. We are currently studying tools for automating generation and verification of rules starting from a high level behavioral representation.
}

\bibliographystyle{IEEEtran}
\bibliography{portmonitor_application}

\end{document}